\documentclass[runningheads,a4paper]{llncs}

\usepackage{url}
\usepackage{amssymb}
\usepackage[norelsize,linesnumbered,ruled,vlined]{algorithm2e}
\usepackage{amsmath}
\usepackage{array}
\newcolumntype{P}[1]{>{\centering\arraybackslash}p{#1}}
\usepackage{graphicx}
\usepackage[export]{adjustbox}
\usepackage{algorithmic}

\urldef{\mailsa}\path| majdi.khaled@gmail.com,indrakshi.ray@colostate.edu,|
\urldef{\mailsb}\path|chitsaz@chitsazlab.org|
\newcommand{\keywords}[1]{\par\addvspace\baselineskip
\noindent\keywordname\enspace\ignorespaces#1}

\begin{document}

\mainmatter

\title{Confidence-Weighted Bipartite Ranking}
\titlerunning{Confidence-Weighted Bipartite Ranking}  
%
\author{Majdi Khalid \and Indrakshi Ray \and Hamidreza Chitsaz}

\institute{Colorado State University, Fort Collins, USA\\
\mailsa\\
\mailsb\\
\url{www.colostate.edu}
}

\toctitle{Confidence-Weighted Bipartite Ranking}
\maketitle              

\begin{abstract}
Bipartite ranking is a fundamental machine learning and data mining problem. It commonly concerns the maximization of the AUC metric. Recently, a number of studies have proposed online bipartite ranking algorithms to learn from massive streams of class-imbalanced data. These methods suggest both linear and kernel-based bipartite ranking algorithms based on first and second-order online learning. Unlike kernelized ranker, linear ranker is more scalable learning algorithm. The existing linear online bipartite ranking algorithms lack either handling non-separable data or constructing adaptive large margin. These limitations yield unreliable bipartite ranking performance. In this work, we propose a linear online confidence-weighted bipartite ranking algorithm (CBR) that adopts soft confidence-weighted learning. The proposed algorithm leverages the same properties of soft confidence-weighted learning in a framework for bipartite ranking. We also develop a diagonal variation of the proposed confidence-weighted bipartite ranking algorithm to deal with high-dimensional data by maintaining only the diagonal elements of the covariance matrix. We empirically evaluate the effectiveness of the proposed algorithms on several benchmark and high-dimensional datasets. The experimental results validate the reliability of the proposed algorithms. The results also show that our algorithms outperform or are at least comparable to the competing online AUC maximization methods.
\end{abstract}

\section{Introduction}

Bipartite ranking is a fundamental machine learning and data mining problem because of its wide range of applications such as recommender systems, information retrieval, and bioinformatics \cite{agarwal2005study,liu2009learning,rendle2009learning}. Bipartite ranking has also been shown to be an appropriate learning algorithm for imbalanced data \cite{cortes2004auc}. The aim of the bipartite ranking algorithm is to maximize the Area Under the Curve (AUC) by learning a function that scores positive instances higher than negative instances. Therefore, the optimization problem of such a ranking model is formulated as the minimization of a pairwise loss function. This ranking problem can be solved by applying a binary classifier to pairs of positive and negative instances, where the classification function learns to classify a pair as positive or negative based on the first instance in the pair. The key problem of this approach is the high complexity of a learning algorithm that grows quadratically or subquadratically \cite{joachims2006training} with respect to the number of instances.

Recently, significant efforts have been devoted to developing scalable bipartite ranking algorithms to optimize AUC in both batch and online settings \cite{ding2015adaptive,gao2013one,hu2015kernelized,li2014top,zhao2011online}. Online learning approach is an appealing statistical method because of its scalability and effectivity. The online bipartite ranking algorithms can be classified on the basis of the learned function into linear and nonlinear ranking models. While they are advantageous over linear online ranker in modeling nonlinearity of the data, kernelized online ranking algorithms require a kernel computation for each new training instance. Further, the decision function of the nonlinear kernel ranker depends on support vectors to construct the kernel and to make a decision.

Online bipartite ranking algorithms can also be grouped into two different schemes. The first scheme maintains random instances from each class label in a finite buffer \cite{kar2013generalization,zhao2011online}. Once a new instance is received, the buffer is updated based on stream oblivious policies such as Reservoir Sampling (RS) and First-In-First-Out (FIFO). Then the ranker function is updated based on a pair of instances, the new instance and each opposite instance stored in the corresponding buffer. These online algorithms are able to deal with non-separable data, but they are based on simple first-order learning. The second approach maintains the first and second statistics \cite{gao2013one}, and is able to adapt the ranker to the importance of the features \cite{ding2015adaptive}. However, these algorithms assume the data are linearly separable.

Moreover, these algorithms make no attempt to exploit the confidence information, which has shown to be very effective in ameliorating the classification performance \cite{crammer2009multi,dredze2008confidence,wang2012exact}. Confidence-weighted (CW) learning takes the advantage of the underlying structure between features by modeling the classifier (i.e., the weight vector) as a Gaussian distribution parameterized by a mean vector and covariance matrix \cite{dredze2008confidence}. This model captures the notion of confidence for each weight coordinate via the covariance matrix. A large diagonal value corresponding to the i-th feature in the covariance matrix results in less confidence in its weight (i.e., its mean). Therefore, an aggressive update is performed on the less confident weight coordinates. This is analogous to the adaptive subgradient method \cite{duchi2011adaptive} that involves the geometric structure of the data seen so far in regularizing the weights of sparse features (i.e., less occurring features) as they are deemed more informative than dense features. The confidence-weighted algorithm \cite{dredze2008confidence} has also been improved by introducing the adaptive regularization (AROW) that deals with inseparable data \cite{crammer2009adaptive}. The soft confidence-weighted (SCW) algorithm improves upon AROW by maintaining an adaptive margin \cite{wang2012exact}.

In this paper, we propose a novel framework that solves a linear online bipartite ranking using the soft confidence-weighted algorithm \cite{wang2012exact}. The proposed confidence-weighted bipartite ranking algorithm (CBR) entertains the fast training and testing phases of linear bipartite ranking. It also enjoys the capability of the soft confidence-weighted algorithm in learning confidence-weighted model, handling linearly inseparable data, and constructing adaptive large margin. The proposed framework follows an online bipartite ranking scheme that maintains a finite buffer for each class label while updating the buffer by one of the stream oblivious policies such as Reservoir Sampling (RS) and First-In-First-Out (FIFO) \cite{vitter1985random}. We also provide a diagonal variation (CBR-diag) of the proposed algorithm to handle high-dimensional datasets.

The remainder of the paper is organized as follows. In Section 2, we briefly review closely related work. We present the confidence-weighted bipartite ranking (CBR) and its diagonal variation (CBR-diag) in Section 3. The experimental results are presented in Section 4. Section 5 concludes the paper and presents some future work.

\section{Related Work}

The proposed bipartite ranking algorithm is closely related to the online learning and bipartite ranking algorithms. What follows is a brief review of recent studies related to these topics.

\textbf{Online Learning}. The proliferation of big data and massive streams of data emphasize the importance of online learning algorithms. Online learning algorithms have shown a comparable classification performance compared to batch learning, while being more scalable. Some online learning algorithms, such as the Perceptron algorithm \cite{rosenblatt1958perceptron}, the Passive-Aggressive (PA) \cite{crammer2006online}, and the online gradient descent \cite{zinkevich2003online}, update the model based on a first-order learning approach. These methods do not take into account the underlying structure of the data during learning. This limitation is addressed by exploring second-order information to exploit the underlying structure between features in ameliorating learning performance \cite{cesa2005second,crammer2009adaptive,dredze2008confidence,duchi2011adaptive,orabona2010new,wang2012exact}. Moreover, kernelized online learning methods have been proposed to deal with nonlinearly distributed data \cite{cavallanti2007tracking,dekel2008forgetron,orabona2009bounded}.
  
\textbf{Bipartite Ranking}. Bipartite ranking learns a real-valued function that induces an order on the data in which the positive instances precede negative instances. The common measure used to evaluate the success of the bipartite ranking algorithm is the AUC \cite{hanley1982meaning}. Hence the minimization of the bipartite ranking loss function is equivalent to the maximization of the AUC metric. The AUC presents the probability that a model will rank a randomly drawn positive instance higher than a random negative instance. In batch setting, a considerable amount of studies have investigated the optimization of linear and nonlinear kernel ranking function \cite{chapelle2010efficient,joachims2005support,kuo2014large,lee2014large}. More scalable methods are based on stochastic or online learning \cite{sculley2009large,wan2015online,wangsolar}. However, these methods are not specifically designed to optimize the AUC metric. Recently, a few studies have focused on the optimization of the AUC metric in online setting. The first approach adopted a framework that maintained a buffer with limited capacity to store random instances to deal with the pairwise loss function \cite{kar2013generalization,zhao2011online}. The other methods maintained only the first and second statistics for each received instance and optimized the AUC in one pass over the training data \cite{ding2015adaptive,gao2013one}. The work \cite{ding2015adaptive} exploited the second-order technique \cite{duchi2011adaptive} to make the ranker aware of the importance of less frequently occurring features, hence updating their weights with a higher learning rate.

Our proposed method follows the online framework that maintains fixed-sized buffers to store instances from each class label. Further, it exploits the online second-order method \cite{wang2012exact} to learn a robust bipartite ranking function. This distinguishes the proposed method from \cite{kar2013generalization,zhao2011online} employing first-order online learning. Also, the proposed method is different from \cite{ding2015adaptive,gao2013one} by learning a confidence-weighted ranker capable of dealing with non-separable data and learning an adaptive large margin. The most similar approaches to our method are \cite{wan2015online,wangsolar}. However, these are not designed to directly maximize the AUC. They also use classical first-order and second-order online learning whereas we use the soft variation of confidence-weighted learning that has shown a robust performance in the classification task \cite{wang2012exact}.

\section{Online Confidence-Weighted Bipartite Ranking}

\subsection{Problem Setting}

We consider a linear online bipartite ranking function that learns on imbalanced data to optimize the AUC metric \cite{hanley1982meaning}. Let $\mathcal{S} = \{x_{i}^{+} \cup x_{j}^{-}\ \in \mathcal{R}^{d} | i = \{1,\ldots,n\} , j = \{ 1, \ldots, m\} \}$ denotes the input space of dimension $d$ generated from unknown distribution $\mathcal{D}$, where $x_{i}^{+}$ is the i-th positive instance and $x_{j}^{-}$ is the j-th negative instance. The $n$ and $m$ denote the number of positive and negative instances, respectively. The linear bipartite ranking function $f:\mathcal{S} \rightarrow \mathcal{R}$ is a real valued function that maximizes the AUC metric by minimizing the following loss function:

\begin{eqnarray*}\label{eq1}
\mathcal{L}(f;\mathcal{S}) = \frac{1}{nm} \sum_{i=1}^{n} \sum_{j=1}^{m}  I(f(x_{i}^{+}) \leq f(x_{j}^{-})),
\end{eqnarray*}

where $f(x) = w^{T}x$ and $I(\cdot)$ is an indicator function that outputs  $1$ if the condition is held, and $0$ otherwise. It is common to replace the indicator function with a convex surrogate function,

\begin{eqnarray}\label{eq2}
\mathcal{L}(f;\mathcal{S}) = \frac{1}{nm} \sum_{i=1}^{n} \sum_{j=1}^{m} \ell(f(x_{i}^{+}) - f(x_{j}^{-}) ),
\end{eqnarray}

\noindent
where $\ell(\cdot)$ is a surrogate loss function such as hinge loss $\ell(z) = max(0,1-z)$.

It is easy to see that the complexity of optimizing (\ref{eq2}) will grow quadratically with respect to the number of training instances. Following the approach suggested by \cite{zhao2011online} to deal with the complexity of the pairwise loss function, we reformulate the pairwise loss function (\ref{eq2}) as a sum of two losses for a pair of instances,

\begin{eqnarray} \label{eq4}
 \sum_{t=1}^{T} I_{(y_{t} = +1)} g_{t}^{+} (w) + I_{(y_{t} = -1)} g_{t}^{-} (w),
\end{eqnarray}

where $T = n^{+}+n^{-}$, and $g_{t}(w)$ is defined as follows

\begin{eqnarray}\label{1eq5}
 g_{t}^{+}(w) =  \sum_{t'=1}^{t-1} I_{(y_{t'} = -1)}  \ell(f(x_{t}) - f(x_{t'})),
\end{eqnarray}

\begin{eqnarray}\label{eq6}
 g_{t}^{-}(w) =  \sum_{t'=1}^{t-1} I_{(y_{t'} = +1)}  \ell(f(x_{t'}) - f(x_{t})) .
\end{eqnarray}

Instead of maintaining all the received instances to compute the gradients $\nabla g_{t}(w)$, we store random instances from each class in the corresponding buffer. Therefore, two buffers $B_{+}$ and $B_{-}$ with predefined capacity are maintained for positive and negative classes, respectively. The buffers are updated using a stream oblivious policy. The current stored instances in a buffer are used to update the classifier as in equation (\ref{eq4}) whenever a new instance from the opposite class label is received. 

The framework of the online confidence-weighted bipartite ranking is shown in Algorithm \ref{alg1}. The two main components of this framework are UpdateBuffer and UpdateRanker, which are explained in the following subsections.

\begin{algorithm}[t]
  \caption{A Framework for Confidence-Weighted Bipartite Ranking (CBR)} \label{alg1}
   \begin{algorithmic}
       \STATE {\bf Input}: \begin{itemize}

       \item[\textbullet] the penalty parameter $C$ \\
       \item[\textbullet] $ $the capacity of the buffers $M_{+}$ and $M_{-}$ \\
 \item[\textbullet] $ \eta$ parameter \\
 \item[\textbullet] $ a_{i} = 1$ for $i \in {1,\dots,d}$
\end{itemize}

\STATE {\bf Initialize}: $\mu_{1} = \{0,\dots,0\}^{d}$, $B_{+} = B_{-} = \emptyset$, $M^{1}_{+} = M^{1}_{-} = 0 $\\
$\qquad \qquad \quad \Sigma_{1} = diag(a)$ or $G_{1} = a $
\FOR{$t = 1, \ldots, T$}

\STATE Receive a training instance $(x_{t},y_{t})$
\IF{$y_{t} = +1$}
\STATE	$B_{-}^{t+1} = B_{-}^{t}$, $M^{t+1}_{+}= M^{t}_{+} + 1$, $M^{t+1}_{-} = M^{t}_{-}$ \\
\STATE	 $C_{t} = C$ 
\STATE	 $B_{+}^{t+1} = $ UpdateBuffer($x_{t},B_{+}^{t},M_{+}$,$M^{t+1}_{+}$) \\
\STATE $[ \mu_{t+1},\Sigma_{t+1}] =$ UpdateRanker($\mu_{t},\Sigma_{t},x_{t},y_{t},C_{t},B_{-}^{t+1}, \eta$) or \\
\STATE $[ \mu_{t+1},G_{t+1}] =$ UpdateRanker($\mu_{t},G_{t},x_{t},y_{t},C_{t},B_{-}^{t+1}, \eta$) (CBR-diag)

\ELSE
\STATE $B_{+}^{t+1} = B_{+}^{t}$,$M^{t+1}_{-}= M^{t}_{-} + 1$, $M^{t+1}_{+} = M^{t}_{+}$ \\
\STATE $C_{t} = C$  
\STATE  $B_{-}^{t+1} = $ UpdateBuffer($x_{t},B_{-}^{t},M_{-}$,$M^{t+1}_{-}$) \\
\STATE $[\mu_{t+1},\Sigma_{t+1}]=$ UpdateRanker($\mu_{t},\Sigma_{t},x_{t},y_{t},C_{t},B_{+}^{t+1},\eta$) or \\
\STATE $[\mu_{t+1},G_{t+1}]=$ UpdateRanker($\mu_{t},G_{t},x_{t},y_{t},C_{t},B_{+}^{t+1},\eta$) (CBR-diag)

\ENDIF
\ENDFOR
\end{algorithmic}
\end{algorithm}

\subsection{Update Buffer}
One effective approach to deal with pairwise learning algorithms is to maintain a buffer with a fixed capacity. This raises the problem of updating the buffer to store the most informative instances. In our online Bipartite ranking framework, we investigate the following two stream oblivious policies to update the buffer:

Reservoir Sampling \textbf{(RS)}: Reservoir Sampling is a common oblivious policy to deal with streaming data \cite{vitter1985random}. In this approach, the new instance $(x_{t},y_{t})$ is added to the corresponding buffer if its capacity is not reached, $|B_{y_{t}}^{t}| < M_{y_{t}}$. If the buffer is at capacity, it will be updated with probability $\frac{M_{y_{t}}}{M^{t+1}_{y_{t}}}$ by randomly replacing one instance in $B^{t}_{y_{t}}$ with $x_{t}$. Algorithm \ref{resersam} shows the steps of the Reservoir sampling approach for updating the buffers.

First-In-First-Out \textbf{(FIFO)}: This simple strategy replaces the oldest instance with the new instance if the corresponding buffer reaches its capacity. Otherwise, the new instance is simply added to the buffer.

\begin{algorithm}[t]
  \caption{Reservoir Sampling Approach} \label{resersam}
   \begin{algorithmic}
       \STATE {\bf Input}: $x_{t}$,$\;B^{t}$,$\;M$,$\;M_{t+1}$\\ 
       
       \STATE {\bf Output}: updated buffer $B^{t+1}$\\ 
        
        \IF{$|B^{t}| < M$} 
        \STATE $B^{t+1} = B^{t} \cup \{x_{t}\}$
        
        \ELSE
        \STATE Sample $Z$ from a Bernoulli distribution with $Pr(Z = 1) = M/M_{t+1}$ \\
        
        \IF{$Z=1$}
         \STATE Randomly delete an instance from $B^{t}$ \\
         \STATE $B^{t+1} = B^{t} \cup \{x_{t}\}$ \\
        \ENDIF
        \ENDIF
        Return $B^{t+1}$

\end{algorithmic}
\end{algorithm}

\subsection{Update Ranker} 
Inspired by the robust performance of second-order learning algorithms, we apply the soft confidence-weighted learning approach \cite{wang2012exact} to updated the bipartite ranking function. Therefore, our confidence-weighted bipartite ranking model (CBR) is formulated as a ranker with a Gaussian distribution parameterized by mean vector $\mu \in  \mathcal{R}^{d}$ and covariance matrix $\Sigma \in  \mathcal{R}^{d \times d}$. The mean vector $\mu$ represents the model of the bipartite ranking function, while the covariance matrix captures the confidence in the model. The ranker is more confident about the model value $\mu_{p}$ as its diagonal value $\Sigma_{p,p}$ is smaller. The model distribution is updated once the new instance is received while being close to the old model distribution. This optimization problem is performed by minimizing the Kullback-Leibler divergence between the new and the old distributions of the model. The online confidence-weighted bipartite ranking (CBR) is formulated as follows:

\begin{eqnarray} \label{eq7}
(\mu_{t+1} , \Sigma_{t+1}) =  \underset{\mu,\Sigma}{\operatorname{argmin}} D_{KL} (\mathcal{N}(\mu,\Sigma) || \mathcal{N}(\mu_{t},\Sigma_{t})) \\
+ C \ell^{\phi}(\mathcal{N}(\mu,\Sigma); (z,y_{t})), \nonumber{}
\end{eqnarray}

where $z = (x_{t}-x)$, $C$ is the the penalty hyperparamter, $\phi = \Phi^{-1}(\eta)$, and $\Phi$ is the normal cumulative distribution function. The loss function $\ell^{\phi}(\cdot)$ is defined as:

\begin{eqnarray*}\label{eq8}
\ell^{\phi}(\mathcal{N}(\mu,\Sigma); (z,y_{t})) = max(0,\phi \sqrt{z^{T}\Sigma z} - y_{t} \mu \cdot z).
\end{eqnarray*}

The solution of (\ref{eq7}) is given by the following proposition.

\begin{proposition}\label{prop1}
 The optimization problem (\ref{eq7}) has a closed-form solution as follows:
\begin{equation*}
\mu_{t+1} = \mu_{t} + \alpha_{t}y_{t}\Sigma_{t}z,
\end{equation*}
\begin{equation*}
\Sigma_{t+1} = \Sigma_{t}  - \beta_{t} \Sigma_{t} z^{T} z\Sigma_{t}.
\end{equation*}

The coefficients $\alpha$ and $\beta$ are defined as follows: \\

$\alpha_{t} = min\{ C, max \{ 0,\frac{1}{\upsilon_{t} \zeta} (-m_{t} \psi + \sqrt{m_{t}^{2} \frac{\phi^{4}}{4} + \upsilon_{t} \phi^{2} \zeta } ) \} \}$, \\

$\beta_{t} = \frac{\alpha_{t} \phi}{ \sqrt{u_{t}} + \upsilon_{t} \alpha_{t} \phi}$.
where $u_{t} = \frac{1}{4} ( - \alpha_{t} \upsilon_{t} \phi + \sqrt{\alpha_{t}^{2} \upsilon_{t}^{2} \phi^{2} + 4 \upsilon_{t} })^{2}$, 

$\upsilon_{t} = z^{T} \Sigma_{t} z$, $m_{t} = y_{t}(\mu_{t} \cdot z)$ , $ \phi  = \Phi^{-1}(\eta)$, $ \psi = 1 + \frac{\phi^{2}}{2}$, $\zeta = 1 + \phi^{2} $, and 
 
 $z = x_{t} - x$.
\end{proposition}

\noindent The proposition (\ref{prop1}) is analogous to the one derived in \cite{wang2012exact}.\\

Though modeling the full covariance matrix lends the CW algorithms a powerful capability in learning \cite{crammer2012confidence,ma2010exploiting,wang2012exact}, it raises potential concerns with high-dimensional data. The covariance matrix grows quadratically with respect to the data dimension. This makes the CBR algorithm impractical with high-dimensional data due to high computational and memory requirements. 

We remedy this deficiency by a diagonalization technique \cite{crammer2012confidence,duchi2011adaptive}. Therefore, we present a diagonal confidence-weighted bipartite ranking (CBR-diag) that models the ranker as a mean vector $\mu \in  \mathcal{R}^{d}$ and diagonal matrix $\hat{\Sigma} \in  \mathcal{R}^{d \times d}$. Let $G$ denotes $diag(\hat{\Sigma})$, and the optimization problem of CBR-diag is formulated as follows:

\begin{eqnarray} \label{eq9}
(\mu_{t+1} , G_{t+1}) =  \underset{\mu,G}{\operatorname{argmin}} D_{KL} (\mathcal{N}(\mu,G) || \mathcal{N}(\mu_{t},G_{t})) \\
+ C \ell^{\phi}(\mathcal{N}(\mu,G); (z,y_{t})). \nonumber{}
\end{eqnarray}

\begin{proposition}\label{prop2} 
The optimization problem (\ref{eq9}) has a closed-form solution as follows:\\
\begin{equation*}
\mu_{t+1} = \mu_{t} + \frac{\alpha_{t}y_{t}z}{G_{t}},
\end{equation*}
\begin{equation*}
G_{t+1} = G_{t}  + \beta_{t} z^{2}.
\end{equation*}

The coefficients $\alpha$ and $\beta$ are defined as follows \\

$\alpha_{t} = min\{ C, max \{ 0,\frac{1}{\upsilon_{t} \zeta} (-m_{t} \psi + \sqrt{m_{t}^{2} \frac{\phi^{4}}{4} + \upsilon_{t} \phi^{2} \zeta } ) \} \}$, \\

$ \beta_{t} = \frac{\alpha_{t} \phi}{ \sqrt{u_{t}} + \upsilon_{t} \alpha_{t} \phi}$,

where $u_{t} = \frac{1}{4} ( - \alpha_{t} \upsilon_{t} \phi + \sqrt{\alpha_{t}^{2} \upsilon_{t}^{2} \phi^{2} + 4 \upsilon_{t} })^{2}$, $\upsilon_{t} = \sum_{i=1}^{d} \frac{z_{i}^{2}}{G_{i}+C}$,  $m_{t} = y_{t}(\mu_{t} \cdot z)$, 

$\phi  = \Phi^{-1}(\eta)$, $\psi = 1 + \frac{\phi^{2}}{2}$, $\zeta = 1 + \phi^{2}$, and $z = x_{t} - x$.

\end{proposition}
The propositions \ref{prop1} and \ref{prop2} can be proved similarly to the proof in \cite{wang2012exact}. The steps of updating the online confidence-weighted bipartite ranking with full covariance matrix or with the diagonal elements are summarized in Algorithm \ref{alg2}.

\begin{algorithm}
\caption{Update Ranker} \label{alg2}
 \begin{algorithmic}
 
\STATE {\bf Input}:
  \begin{itemize}
  \item[\textbullet] $ \mu_{t}\;\;\;\;\;\;\;\;:$ current mean vector \\
   \item[\textbullet] $ \Sigma_{t} \; or \; G_{t}:$ current covariance matrix or diagonal elements \\
   \item[\textbullet] $(x_{t},y_{t}):$ a training instance \\
   \item[\textbullet] $ B\;\;\;\;\;\;\;\;:$ the buffer storing instances from the opposite class label \\
   \item[\textbullet] $ C_{t}\;\;\;\;\;\;\;:$ class-specific weighting parameter \\
   \item[\textbullet] $ \eta\;\;\;\;\;\;\;\;\;\;$: the predefined probability
  \end{itemize}
\STATE {\bf Output}: updated ranker:  
\begin{itemize} 
\item[\textbullet] $\mu_{t+1}$ 
\item[\textbullet] $\Sigma_{t+1} \; or \; G_{t+1}$
\end{itemize}

\STATE{\bf Initialize}: $\mu^{1} = \mu_{t}$, $(\Sigma^{1} = \Sigma_{t}$ or  $G^{1} = G_{t})$, $i=1$

\FOR{$x \in B$}
\STATE Update the ranker $(\mu^{i},\Sigma^{i})$ with $z = x_{t}-x$ and $y_{t}$ by 

\STATE $(\mu^{i+1} , \Sigma^{i+1}) =  \underset{\mu,\Sigma}{\operatorname{argmin}} D_{KL} (\mathcal{N}(\mu,\Sigma) || \mathcal{N}(\mu^{i},\Sigma^{i})) + C \ell^{\phi}(\mathcal{N}(\mu,\Sigma); (z,y_{t}))$

\STATE or \\
\STATE Update the ranker $(\mu^{i},G^{i})$ with $z = x_{t}-x$ and $y_{t}$ by \\

\STATE $ (\mu^{i+1} , G^{i+1}) =  \underset{\mu,G}{\operatorname{argmin}} D_{KL} (\mathcal{N}(\mu,G) || \mathcal{N}(\mu^{i},G^{i})) + C \ell^{\phi}(\mathcal{N}(\mu,G); (z,y_{t}))$ \\

\STATE $i = i + 1$

\ENDFOR
\STATE Return $\mu_{t+1} = \mu^{|B|+1}$ 
\STATE $\qquad \quad \Sigma_{t+1} = \Sigma^{{|B|+1}}$ or $G_{t+1} = G^{{|B|+1}}$
\end{algorithmic}
\end{algorithm}

\section{Experimental Results}
In this section, we conduct extensive experiments on several real world datasets in order to demonstrate the effectiveness of the proposed algorithms. We also compare the performance of our methods with existing online learning algorithms in terms of AUC and classification accuracy at the optimal operating point of the ROC curve (OPTROC). The running time comparison is also presented.

\subsection{Real World Datasets}

We conduct extensive experiments on various benchmark and high-dimensional datasets. All datasets can be downloaded from LibSVM\footnote{https://www.csie.ntu.edu.tw/~cjlin/libsvmtools/} and the machine learning repository UCI\footnote{https://archive.ics.uci.edu/ml/} except the Reuters\footnote{http://www.cad.zju.edu.cn/home/dengcai/Data/TextData.html} dataset that is used in \cite{cai2011locally}. If the data are provided as training and test sets, we combine them together in one set. For cod-rna data, only the training and validation sets are grouped together. For rcv1 and news20, we only use their training sets in our experiments. The multi-class datasets are transformed randomly into class-imbalanced binary datasets. Tables \ref{table1} and \ref{table2} show the characteristics of the benchmark and the high-dimensional datasets, respectively.

\begin{table*}
\caption{Benchmark datasets}
\label{table1}
\centering

\begin{tabular}{|c|c c|c|c c|c|c c|} \hline 

Data      &   \#inst  &   \#feat    &  Data                &   \#inst    &  \#feat      & Data                &   \#inst   &   \#feat \\
 \hline

glass     & 214 & 10                  & cod-rna   & 331,152  & 8                     & australian          & 690     &  14 \\ \hline

ionosphere & 351   & 34           & spambase &  4,601  & 57                    & diabetes          & 768     &  8 \\ \hline

german  & 1,000   &  24           & covtype   & 581,012  &  54                   & acoustic          & 78,823     &  50 \\ \hline
 
svmguide4 & 612  & 10            & magic04  &  19,020   &  11                  & vehicle          & 846     &  18 \\ \hline

svmguide3 & 1284  & 21          & heart   &  270   &  13                            & segment          & 2,310     &  19 \\ \hline

\end{tabular}

\end{table*}

\begin{table*}
\caption{ High-dimensional datasets}
\label{table2}
\centering
\begin{tabular}{ |P{1.8cm}|P{1.2cm} P{1.2cm}| } \hline

Data      &   \#inst  &   \#feat   \\ \hline

farm-ads  & 4,143   &  54,877   \\ \hline

rcv1    &  15,564  &  47,236  \\ \hline

sector &  9,619  &  55,197   \\ \hline

real-sim & 72,309  &  20,958   \\ \hline

news20 & 15,937  &  62,061   \\ \hline

Reuters &  8,293  &  18,933  \\ \hline

\end{tabular}
\end{table*}

\subsection{Compared Methods and Model Selection}

\textbf{Online Uni-Exp} \cite{kotlowski2011bipartite}: An online pointwise ranking algorithm that optimizes the weighted univariate exponential loss. The learning rate is tuned by 3-fold cross validation on the training set by searching in $2^{[-10:10]}$.

\noindent  \textbf{OPAUC} \cite{gao2013one}: An online learning algorithm that optimizes the AUC in one-pass through square loss function. The learning rate is tuned by 3-fold cross validation by searching in $2^{[-10:10]}$, and the regularization hyperparameter is set to a small value 0.0001.

\noindent  \textbf{OPAUCr} \cite{gao2013one}: A variation of OPAUC that approximates the covariance matrices using low-rank matrices. The model selection step is carried out similarly to OPAUC, while the value of rank $\tau$ is set to 50 as suggested in  \cite{gao2013one}.

\noindent \textbf{OAM$_{\text{seq}}$} \cite{zhao2011online}: The online AUC maximization (OAM) is the state-of-the-art first-order learning method. We implement the algorithm with the Reservoir Sampling as a buffer updating scheme. The size of the positive and negative buffers is fixed at 50. The penalty hyperparameter $C$ is tuned by 3-fold cross validation on the training set by searching in $2^{[-10:10]}$.

\noindent \textbf{AdaOAM} \cite{ding2015adaptive}: This is a second-order AUC maximization method that adapts the classifier to the importance of features. The smooth hyperparameter $\delta$ is set to 0.5, and the regularization hyperparameter is set to 0.0001. The learning rate is tuned by 3-fold cross validation on the training set by searching in $2^{[-10:10]}$. 

\noindent \textbf{CBR$_{\text{RS}}$} and \textbf{CBR$_{\text{FIFO}}$}: The proposed confidence-weighted bipartite ranking algorithms with the Reservoir Sampling and First-In-First-Out buffer updating policies, respectively. The size of the positive and negative buffers is fixed at 50. The hyperparameter $\eta$ is set to 0.7, and the penalty hyperparameter $C$ is tuned by 3-fold cross validation by searching in $2^{[-10:10]}$.

\noindent \textbf{CBR-diag$_{\text{FIFO}}$}: The proposed diagonal variation of confidence-weighted bipartite ranking that uses the First-In-First-Out policy to update the buffer. The buffers are set to 50, and the hyperparameters are tuned similarly to CBR$_{\text{FIFO}}$. 

For a fair comparison, the datasets are scaled similarly in all experiments. We randomly divide each dataset into 5 folds, where 4 folds are used for training and one fold is used as a test set. For benchmark datasets, we randomly choose 8000 instances if the data exceeds this size. For high-dimensional datasets, we limit the sample size of the data to 2000 due to the high dimensionality of the data. The results on the benchmark and the high-dimensional datasets are averaged over 10 and 5 runs, respectively. A random permutation is performed on the datasets with each run. All experiments are conducted with Matlab 15 on a workstation computer with 8x2.6G CPU and 32 GB memory.

\subsection{Results on Benchmark Datasets}

\begin{figure}[t!]
\centering
\includegraphics[height=6cm,width=13cm]{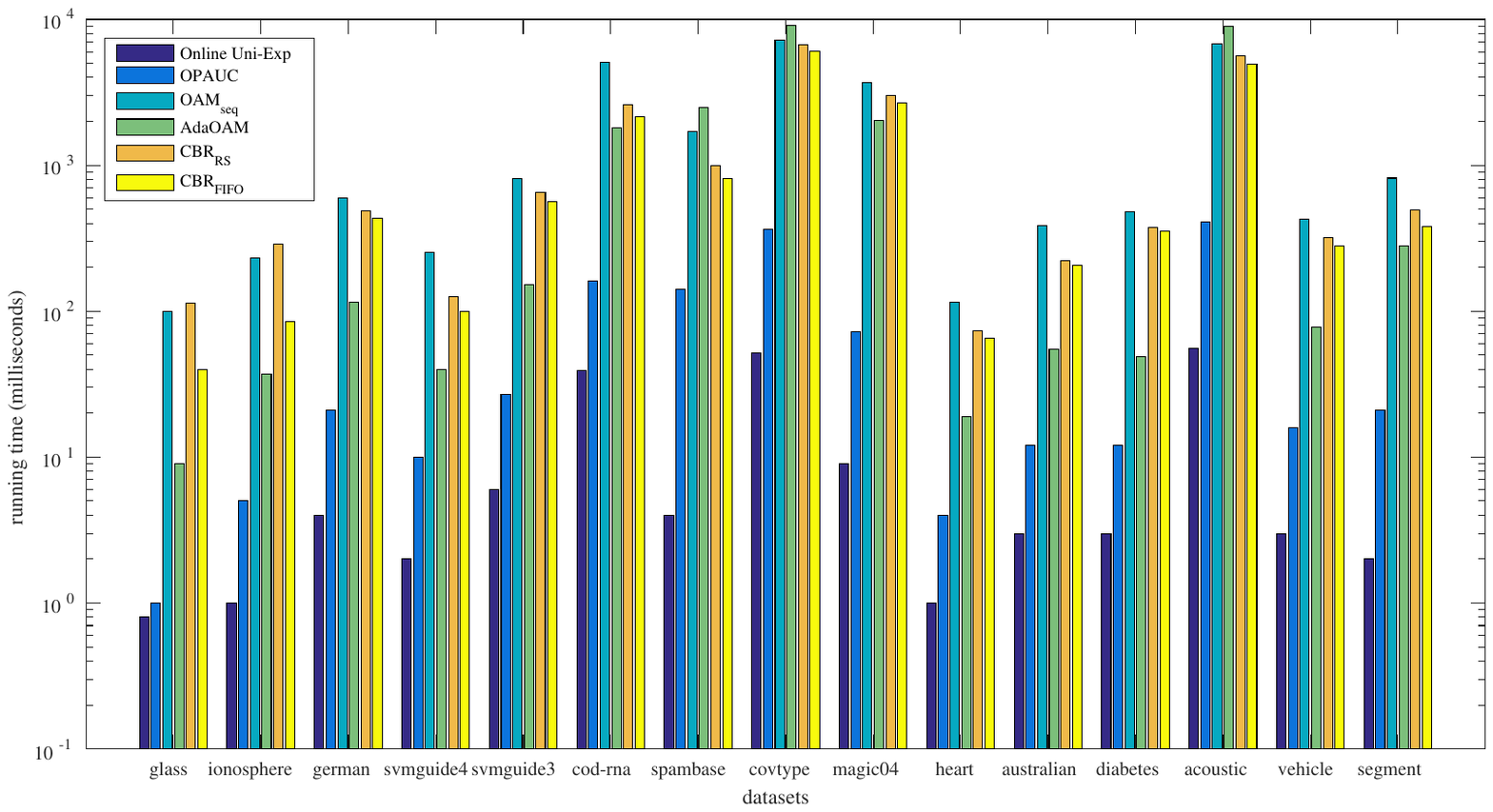}

\caption{Running time (in milliseconds) of CBR and the other online learning algorithms on the benchmark datasets. The \textit{y}-axis is displayed in log- scale.}
\label{fig1}
\end{figure}

The comparison in terms of AUC is shown in Table \ref{table3}, while the comparison in terms of classification accuracy at OPTROC is shown in Table \ref{table4}. The running time (in milliseconds) comparison is illustrated in Figure \ref{fig1}.

The results show the robust performance of the proposed methods CBR$_{RS}$ and CBR$_{FIFO}$ in terms of AUC and classification accuracy compared to other first and second-order online learning algorithms. We can observe that the improvement of the second-order methods such as OPAUC and AdaOAM over first-order method OAM$_{\text{seq}}$ is not reliable, while our CBR algorithms often outperform the OAM$_{\text{seq}}$. Also, the proposed methods are faster than OAM$_{\text{seq}}$, while they incur more running time compared to AdaOAM except with spambase, covtype, and acoustic datasets. The pointwise method online Uni-Exp maintains fastest running time, but at the expense of the AUC and classification accuracy. We also notice that the performance of CBR$_{\text{FIFO}}$ is often slightly better than CBR$_{\text{RS}}$ in terms of AUC, classification accuracy, and running time. 

\begin{table*}
\caption{Comparison of AUC performance on benchmark datasets}
\label{table3}
\centering
\begin{scriptsize}
\begin{tabular}{|c|c|c|c|c|c|c|} \hline

Data    &   CBR$_{\text{RS}}$  & CBR$_{\text{FIFO}}$ &  Online Uni-Exp  &   OPAUC     &  OAM$_{\text{seq}}$      &  AdaOAM  \\ \hline

glass   & \textbf{ 0.825} $\pm$ 0.043 &  0.823 $\pm$ 0.046  &  0.714 $\pm$ 0.075  & 0.798 $\pm$ 0.061 &  0.805 $\pm$ 0.047  & 0.794 $\pm$ 0.061   \\ \hline

ionosphere  & 0.950 $\pm$ 0.027  & \textbf{ 0.951} $\pm$ 0.028 &  0.913 $\pm$ 0.018 &  0.943 $\pm$ 0.026  & 0.946 $\pm$ 0.025  &  0.943 $\pm$ 0.029  \\ \hline

german   &\textbf{ 0.782} $\pm$ 0.024  & 0.780 $\pm$ 0.019  &  0.702 $\pm$ 0.032  & 0.736 $\pm$ 0.034 &  0.731 $\pm$ 0.028   & 0.770 $\pm$ 0.024  \\ \hline

svmguide4  & 0.969 $\pm$ 0.013  & \textbf{0.974} $\pm$ 0.013 &  0.609 $\pm$ 0.096  & 0.733 $\pm$ 0.056 &  0.771 $\pm$ 0.063  & 0.761 $\pm$ 0.053  \\ \hline

svmguide3   & 0.755 $\pm$ 0.022  & \textbf{0.764} $\pm$ 0.036 &  0.701 $\pm$ 0.025 &  0.737 $\pm$ 0.029  & 0.705 $\pm$ 0.033   &0.738 $\pm$ 0.033  \\ \hline

cod-rna   & 0.983 $\pm$ 0.000  & \textbf{0.984} $\pm$ 0.000 &  0.928 $\pm$ 0.000 &  0.927 $\pm$ 0.001 & 0.951 $\pm$ 0.025  & 0.927 $\pm$ 0.000  \\ \hline

spambase  & 0.941 $\pm$ 0.006 &  \textbf{0.942} $\pm$ 0.006 &  0.866 $\pm$ 0.016 &  0.849 $\pm$ 0.020 &  0.897 $\pm$ 0.043  & 0.862 $\pm$ 0.011  \\ \hline

covtype   & 0.816 $\pm$ 0.003  & \textbf{0.835} $\pm$ 0.001  & 0.705 $\pm$ 0.033 &   0.711 $\pm$ 0.041 &  0.737 $\pm$ 0.023  & 0.770 $\pm$ 0.010  \\ \hline

magic04   & 0.799 $\pm$ 0.006 & \textbf{0.801} $\pm$ 0.006 &  0.759 $\pm$ 0.006 &  0.748 $\pm$ 0.033 &  0.757 $\pm$ 0.015 &  0.773 $\pm$ 0.006  \\ \hline

heart   &  0.908 $\pm$ 0.019 &  \textbf{0.909} $\pm$ 0.021 &  0.733 $\pm$ 0.039 &  0.788 $\pm$ 0.054  & 0.806 $\pm$ 0.059 &  0.799 $\pm$ 0.079  \\ \hline

australian   &0.883 $\pm$ 0.028  & \textbf{0.889} $\pm$ 0.019 &  0.710 $\pm$ 0.130  & 0.735 $\pm$ 0.138 &  0.765 $\pm$ 0.107  & 0.801 $\pm$ 0.037  \\ \hline

diabetes    &0.700 $\pm$ 0.021 &  \textbf{0.707} $\pm$ 0.033  & 0.633 $\pm$ 0.036 &  0.667 $\pm$ 0.041  & 0.648 $\pm$ 0.040 &  0.675 $\pm$ 0.034  \\ \hline

acoustic   &0. 879 $\pm$ 0.006 & \textbf{ 0.892} $\pm$ 0.003  & 0.876 $\pm$ 0.003 &  0.878 $\pm$ 0.003 &  0.863 $\pm$ 0.011 &   0.882 $\pm$ 0.003  \\ \hline

vehicle   & \textbf{0.846} $\pm$ 0.031 & \textbf{ 0.846} $\pm$ 0.034 &  0.711 $\pm$ 0.053 &  0.764 $\pm$ 0.073 &  0.761 $\pm$ 0.078 &  0.792 $\pm$ 0.049  \\ \hline

segment  &  0.900 $\pm$ 0.013 &  \textbf{0.903} $\pm$ 0.008  & 0.689 $\pm$ 0.061  & 0.828 $\pm$ 0.024  & 0.812 $\pm$ 0.035  & 0.855 $\pm$ 0.008  \\ \hline

\end{tabular}
\end{scriptsize}
\end{table*}

\begin{table*}
\caption{Comparison of classification accuracy at OPTROC on benchmark datasets}
\label{table4}
\centering
\begin{scriptsize}
\begin{tabular}{|c|c|c|c|c|c|c|} \hline

Data    &   CBR$_{\text{RS}}$  & CBR$_{\text{FIFO}}$ &  Online Uni-Exp  &   OPAUC     &  OAM$_{\text{seq}}$      &  AdaOAM  \\ \hline

glass    & \textbf{0.813} $\pm$ 0.044  & 0.811 $\pm$ 0.049  &  0.732 $\pm$ 0.060 &  0.795 $\pm$ 0.046 &  0.788 $\pm$ 0.040  & 0.783 $\pm$ 0.047 \\ \hline

ionosphere &  \textbf{0.946} $\pm$ 0.028  & \textbf{0.946} $\pm$ 0.022 &  0.902 $\pm$ 0.028  & 0.936 $\pm$ 0.018 &  0.943 $\pm$ 0.017  & 0.938 $\pm$ 0.018 \\ \hline

german   & 0.780 $\pm$ 0.022 &  \textbf{0.787} $\pm$ 0.019  & 0.741 $\pm$ 0.027 &  0.754 $\pm$ 0.022  & 0.751 $\pm$ 0.028 &  0.770 $\pm$ 0.030 \\ \hline

svmguide4 &  0.951 $\pm$ 0.014  & \textbf{0.956} $\pm$ 0.012  & 0.829 $\pm$ 0.021 &  0.843 $\pm$ 0.024 &  0.839 $\pm$ 0.022 &  0.848 $\pm$ 0.020 \\ \hline

svmguide3 &  0.784 $\pm$ 0.015 &  \textbf{0.793} $\pm$ 0.016 &  0.784 $\pm$ 0.019 &  0.777 $\pm$ 0.024 &  0.780 $\pm$ 0.020  & 0.777 $\pm$ 0.024 \\ \hline

cod-rna  &  0.948 $\pm$ 0.002  & \textbf{0.949} $\pm$ 0.000  & 0.887 $\pm$ 0.001  & 0.887 $\pm$ 0.001  & 0.910 $\pm$ 0.019  & 0.887 $\pm$ 0.001 \\ \hline

spambase & \textbf{ 0.899} $\pm$ 0.009 &  0.898 $\pm$ 0.009  &  0.818 $\pm$ 0.019 &  0.795 $\pm$ 0.022 &  0.849 $\pm$ 0.053 &  0.809 $\pm$ 0.014 \\ \hline

covtype  &  0.746 $\pm$ 0.005  & \textbf{0.766} $\pm$ 0.003 &  0.672 $\pm$ 0.018  & 0.674 $\pm$ 0.021  & 0.685 $\pm$ 0.016  & 0.709 $\pm$ 0.008 \\ \hline

magic04  &  0.769 $\pm$ 0.011 &  \textbf{0.771} $\pm$ 0.006 &  0.734 $\pm$ 0.007  & 0.731 $\pm$ 0.015  & 0.736 $\pm$ 0.013 &  0.752 $\pm$ 0.008 \\ \hline

heart   &  \textbf{0.883} $\pm$ 0.032  & 0.875 $\pm$ 0.026  &  0.716 $\pm$ 0.021  & 0.753 $\pm$ 0.038  & 0.777 $\pm$ 0.043  & 0.772 $\pm$ 0.053  \\ \hline

australian  & 0.841 $\pm$ 0.023 & \textbf{ 0.842} $\pm$ 0.022  & 0.711 $\pm$ 0.056 &  0.725 $\pm$ 0.070  & 0.742 $\pm$ 0.064 &  0.768 $\pm$ 0.036  \\ \hline

diabetes &  \textbf{0.714} $\pm$ 0.029 &  0.705 $\pm$ 0.032   & 0.683 $\pm$ 0.037  & 0.692 $\pm$ 0.040  & 0.694 $\pm$ 0.044  & 0.689 $\pm$ 0.040  \\ \hline

acoustic  & 0. 844 $\pm$ 0.005  & \textbf{0.850} $\pm$ 0.003 &  0.840 $\pm$ 0.005  & 0.839 $\pm$ 0.002 &  0.832 $\pm$ 0.005  & 0.841 $\pm$ 0.003  \\ \hline

vehicle  &  \textbf{0.816} $\pm$ 0.018 &  0.814 $\pm$ 0.018  &  0.764 $\pm$ 0.027  & 0.797 $\pm$ 0.014 &  0.790 $\pm$ 0.029  & 0.805 $\pm$ 0.021  \\ \hline

segment  & \textbf{0.838} $\pm$ 0.015 &  0.836 $\pm$ 0.008   & 0.691 $\pm$ 0.031 &  0.768 $\pm$ 0.027  & 0.755 $\pm$ 0.024 &  0.796 $\pm$ 0.014 \\ \hline

\end{tabular}
\end{scriptsize}
\end{table*}

\subsection{Results on High-Dimensional Datasets}

\begin{figure}[t!]
\centering
 \includegraphics[height=8cm,width=10cm]{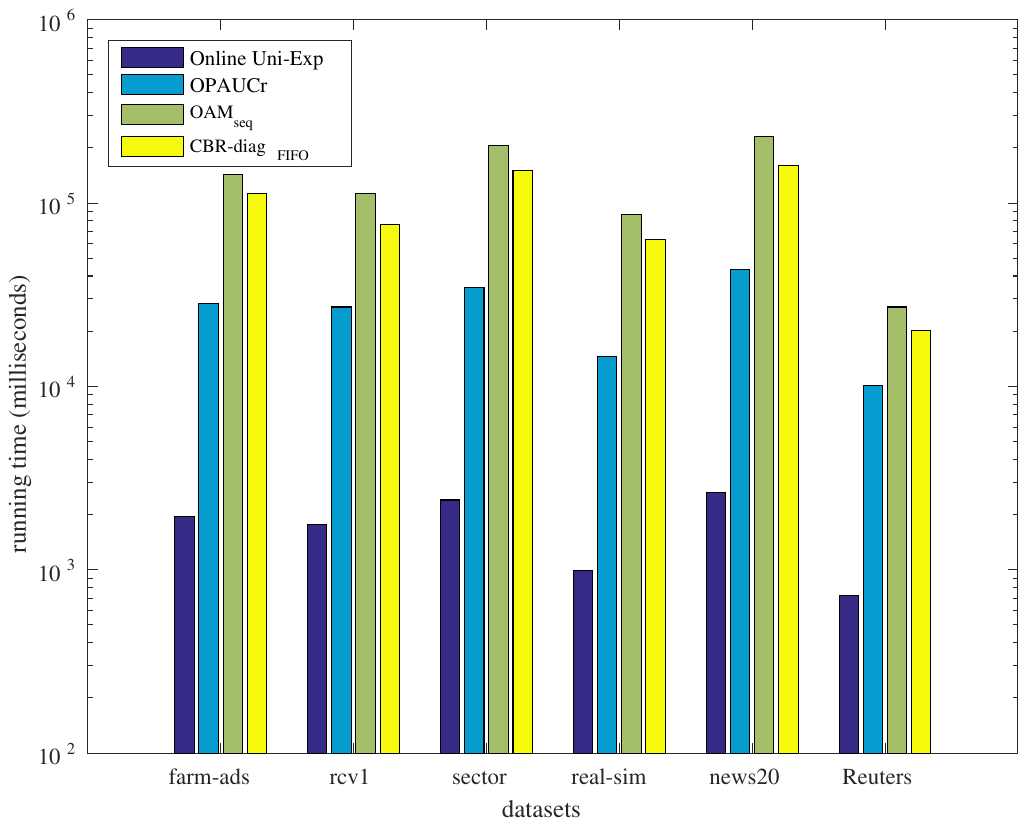}
\caption{Running time (in milliseconds) of CBR-diag$_{FIFO}$ algorithm and the other online learning algorithms on the high-dimensional datasets. The \textit{y}-axis is dis- played in log-scale.}
\label{fig2}
\end{figure}

We study the performance of the proposed CBR-diag$_{FIFO}$ and compare it with online Uni-Exp, OPAUCr, and OAM$_{seq}$ that avoid constructing the full covariance matrix. Table \ref{table5} compares our method and the other online algorithms in terms of AUC, while Table \ref{table6} shows the classification accuracy at OPTROC. Figure \ref{fig2} displays the running time (in milliseconds) comparison.

The results show that the proposed method CBR-diag$_{FIFO}$ yields a better performance on both measures. We observe that the CBR-diag$_{FIFO}$ presents a competitive running time compared to its counterpart OAM$_{seq}$ as shown in Figure \ref{fig2}. We can also see that the CBR-diag$_{FIFO}$ takes more running time compared to the OPAUCr. However, the CBR-diag$_{FIFO}$ achieves better AUC and classification accuracy compared to the OPAUCr. The online Uni-Exp algorithm requires the least running time, but it presents lower AUC and classification accuracy compared to our method.   
 
\begin{table*}[t!]
\caption{Comparison of AUC on high-dimensional datasets}
\label{table5}
\centering
\begin{scriptsize}
\begin{tabular}{|c|c|c|c|c|c|} \hline

Data    &    CBR-diag$_{\text{FIFO}}$ &  Online Uni-Exp  &   OPAUCr     &  OAM$_{\text{seq}}$    \\ \hline

farm-ads  & \textbf{0.961} $\pm$ 0.004  & 0.942 $\pm$ 0.006 &  0.951 $\pm$ 0.004  & 0.952 $\pm$ 0.005 \\ \hline

rcv1  &  \textbf{0.950} $\pm$ 0.007 &  0.927 $\pm$ 0.015  & 0.914 $\pm$ 0.016  & 0.945 $\pm$ 0.008 \\ \hline

sector &  \textbf{0.927} $\pm$ 0.009 &  0.846 $\pm$ 0.019  & 0.908 $\pm$ 0.013 &  0.857 $\pm$ 0.008 \\ \hline

real-sim & \textbf{ 0.982} $\pm$ 0.001  & 0.969 $\pm$ 0.003 &  0.975 $\pm$ 0.002 & 0.977 $\pm$ 0.001 \\ \hline

news20  &\textbf{ 0.956} $\pm$ 0.003 &  0.939 $\pm$ 0.005  & 0.942 $\pm$ 0.006 &  0.944 $\pm$ 0.005 \\ \hline

Reuters  & \textbf{0.993} $\pm$ 0.001 &  0.985 $\pm$ 0.003 &  0.988 $\pm$ 0.002 &  0.989 $\pm$ 0.003 \\ \hline

\end{tabular}
\end{scriptsize}
\end{table*}

\begin{table*}
\caption{Comparison of classification accuracy at OPTROC on high-dimensional datasets}
\label{table6}
\centering
\begin{scriptsize}
\begin{tabular}{|c|c|c|c|c|c|} \hline

Data    &    CBR-diag$_{\text{FIFO}}$ &  Online Uni-Exp  &   OPAUCr     &  OAM$_{\text{seq}}$    \\ \hline

farm-ads  & \textbf{0.897} $\pm$ 0.007 &  0.872 $\pm$ 0.012 &  0.885 $\pm$ 0.008  & 0.882 $\pm$ 0.007 \\ \hline

rcv1  &  \textbf{0.971} $\pm$ 0.001  & 0.967 $\pm$ 0.002  & 0.966 $\pm$ 0.003  & 0.970 $\pm$ 0.001 \\ \hline

sector &  \textbf{0.850} $\pm$ 0.012 &  0.772 $\pm$ 0.011 &  0.831 $\pm$ 0.015 &  0.776 $\pm$ 0.008 \\ \hline

real-sim  & \textbf{0.939} $\pm$ 0.003 &  0.913 $\pm$ 0.005 &  0.926 $\pm$ 0.002  & 0.929 $\pm$ 0.001 \\ \hline

news20  & \textbf{0.918} $\pm$ 0.005 &  0.895 $\pm$ 0.005 &  0.902 $\pm$ 0.009 &  0.907 $\pm$ 0.006 \\ \hline

Reuters  & \textbf{0.971} $\pm$ 0.004 &  0.953 $\pm$ 0.006  & 0.961 $\pm$ 0.006  & 0.961 $\pm$ 0.006 \\ \hline

\end{tabular}
\end{scriptsize}
\end{table*}

\section{Conclusions and Future Work}
In this paper, we proposed a linear online soft confidence-weighted bipartite ranking algorithm that maximizes the AUC metric via optimizing a pairwise loss function. The complexity of the pairwise loss function is mitigated in our algorithm by employing a finite buffer that is updated using one of the stream oblivious policies. We also develop a diagonal variation of the proposed confidence-weighted bipartite ranking algorithm to deal with high-dimensional data by maintaining only the diagonal elements of the covariance matrix instead of the full covariance matrix. The experimental results on several benchmark and high-dimensional datasets show that our algorithms yield a robust performance. The results also show that the proposed algorithms outperform the first and second-order AUC maximization methods on most of the datasets. As future work, we plan to conduct a theoretical analysis of the proposed method. We also aim to investigate the use of online feature selection  \cite{wang2014online} within our proposed framework to effectively handle high-dimensional data.


\end{document}